\documentclass[times, review, 10pt]{elsarticle}
\usepackage[numbers]{natbib}
\setcitestyle{numbers,square}
\usepackage{amsmath,amssymb,amsfonts}
\usepackage{amsmath}
\usepackage{algorithmic}
\usepackage{graphicx}
\usepackage{textcomp}
\usepackage{algorithm}
\usepackage{setspace}
\usepackage{booktabs}
\usepackage{multirow}
\usepackage{booktabs}
\usepackage{cases}
\usepackage{bm}
\usepackage{bbm}
\usepackage{booktabs}
\usepackage{array}
\usepackage{epstopdf}
\usepackage{adjustbox}
\usepackage{soul, color, xcolor}

\usepackage{float}
\usepackage{wrapfig}
\usepackage[caption=false,font=normalsize,labelfont=sf,textfont=sf]{subfig}
\usepackage{textcomp}
\usepackage{stfloats}
\usepackage{url}
\usepackage{verbatim}
\usepackage{gensymb}
\soulregister{\cite}7 
\soulregister{\ref}7 
\soulregister{\pageref}7 

\makeatletter
\def\@begintheorem#1#2{\trivlist
	\item[\hskip \labelsep{\bfseries #1\ #2}]}
\makeatother

\biboptions{sort&compress}
\journal{Engineering Applications of Artificial Intelligence}

\begin{document}
\begin{sloppypar}
\begin{frontmatter}
\let\WriteBookmarks\relax
\def\floatpagepagefraction{1}
\def\textpagefraction{.001}

\title{Graph-Aggregated Prototype Learning for Remote Sensing Semantic Change Detection}

\author[nwpu]{Zhengyi Xu}
\author[nwpu]{Haoran Wu}
\author[nwpu]{Wen Jiang\corref{cor1}}
\ead{jiangwen@nwpu.edu.cn}
\author[nwpu]{Jie Geng}

\cortext[cor1]{Corresponding author.}

\address[nwpu]{School of Electronics and Information,
Northwestern Polytechnical University,
Xi'an 710129, China}

\fntext[funding]{This work was supported in part by
the National Natural Science Foundation of China
under Grant 62271396, and in part by the Innovation
Foundation for Doctoral Dissertations of Northwestern
Polytechnical University under Grant CX2024066.}

\begin{abstract}
	Semantic change detection (SCD) extends the binary change detection task to provide not only the change locations but also the detailed "from-to" categories in multi-temporal remote sensing data. Such detailed semantic insights into changes offer considerable advantages for a wide array of applications. However, since SCD involves the simultaneous optimization of multiple tasks, the model is prone to negative transfer due to task-specific learning difficulties and conflicting gradient flows. To address this issue, we propose Graph-Aggregated Prototype Learning for Remote Sensing Semantic Change Detection(GAPL-SCD). In this framework, a multi-task joint optimization method is designed to optimize the primary task of semantic segmentation and change detection, along with the auxiliary task of graph aggregation prototype learning. Adaptive weight allocation and gradient rotation methods are used to alleviate the conflict between training tasks and improve multi-task learning capabilities. Specifically, the graph aggregation prototype learning module constructs an interaction graph using high-level features. Prototypes serve as class proxies, enabling category-level domain alignment across time points and reducing interference from irrelevant changes. Additionally, the proposed self-query multi-level feature interaction and bi-temporal feature fusion modules further enhance multi-scale feature representation, improving performance in complex scenes. Experimental results on the SECOND and Landsat-SCD datasets demonstrate that our method achieves state-of-the-art performance, with significant improvements in accuracy and robustness for SCD task.
\end{abstract}


\end{frontmatter}
\section{Introduction}
Remote sensing (RS) has emerged as a critical technology for monitoring Earth's surface changes, facilitated by advancements in spatial and spectral resolutions \cite{9782149, 10043760}. High-resolution satellite imagery significantly improves the precision of land-cover dynamic monitoring \cite{8964578}.

Change detection in remote sensing, a crucial component of intelligent image interpretation, aims to analyze and compare satellite images captured at different times over the same geographical region, identifying changes in land-cover states  \cite{bai2023deep}. This technique is widely applied in flood monitoring \cite{saleh2024dam}, emergency response \cite{wang2025refined}, military surveillance, and urban planning\cite{10443352}. With the rapid advancement of artificial intelligence, deep learning methods have been recognized for their powerful feature representation capabilities \cite{10147800, 10440611} and have shown significant potential in change detection tasks \cite{10478593}. Recent studies have extensively discussed deep learning-based remote sensing change detection, highlighting current advancements and existing challenges \cite{10749979}.

However, current approaches often struggle to distinguish fine-grained change types and adequately capture semantic transitions between land-cover categories, limiting their applicability in complex scenarios \cite{long2024semantic}. To address these limitations, semantic change detection (SCD) \cite{11015343} has emerged as a promising solution. This task extracts change regions and category transition information from multi-temporal remote sensing data, providing not only change information but also detailed land cover categories. Recent advancements in SCD have focused on improving feature representation and spatiotemporal modeling. For instance, SCDNET \cite{PENG2021102465} adopts a dual-branch encoder-decoder structure for multi-scale feature extraction. Bi-SRNet \cite{9721305} utilizes self-attention mechanisms to enhance deep semantic interactions across temporal branches. Recently, Transformer-based architectures have also been explored for semantic change detection tasks \cite{Yuan2022ATS, ZHENG2022228}. Pyramid-SCDFormer \cite{Yuan2022ATS} introduced a transformer-based Siamese network that selectively merges semantic tokens in multi-head self-attention blocks to capture fine-grained features and long-range spatiotemporal dependencies, significantly improving the recognition of small-scale changes and fine edges in complex scenarios. ChangeMask \cite{ZHENG2022228} proposes a deep multi-task encoder-decoder architecture that decouples SCD into temporal-wise semantic segmentation and binary change detection, leveraging semantic-change causal relationships and temporal symmetry to enhance feature representation and spatiotemporal modeling. Nevertheless, these methods typically exhibit limited modeling capabilities regarding spatiotemporal dependencies and face challenges from shallow representations.

Two main challenges persist in semantic change detection: 
\begin{itemize}
	\item Semantic change detection involves simultaneous multi-task optimization, often leading to task conflicts and negative transfer due to differing task difficulties and gradient interference.
	\item Practical scenarios frequently introduce irrelevant changes due to sensor variability, weather conditions, and illumination differences, complicating accurate identification and analysis of genuine changes.
\end{itemize}

To address these challenges, we propose a semantic change detection method guided by graph aggregation prototype learning. This approach adopts a joint multi-task optimization strategy to collaboratively optimize semantic segmentation, change detection, and the auxiliary task of graph aggregation prototype learning.

The main contributions of this paper are summarized below:
\begin{itemize}
	\item We propose a multi-task semantic change detection framework that tackles the challenge of conflicting gradient flows caused by varying task difficulties. We leverages uncertainty-based weight allocation to balance task contributions and employs gradient rotation to eliminate conflicting updates, thereby enhancing training stability and overall performance.
	\item We introduce a graph aggregation prototype learning module that constructs an interaction graph using high-level features and employs prototypes as class proxies for category-level domain alignment across bi-temporal images, thereby mitigating the impact of irrelevant changes and other confounders on the change detection task.
	\item We develop a self-query multi-level feature interaction module and a bi-temporal feature fusion module to further strengthen the representation of multi-scale features, improving the performance of semantic segmentation and change detection in complex scenarios.
\end{itemize}

The rest of this paper is organized as follows. Section II reviews related works. Section III describes the proposed GAPL-SCD method. Experimental results and analysis are presented in Section IV, followed by conclusions in Section V.

\begin{figure*}
	\centering
	\begin{tabular}{@{}c@{}c}
		\includegraphics[width=1\linewidth]{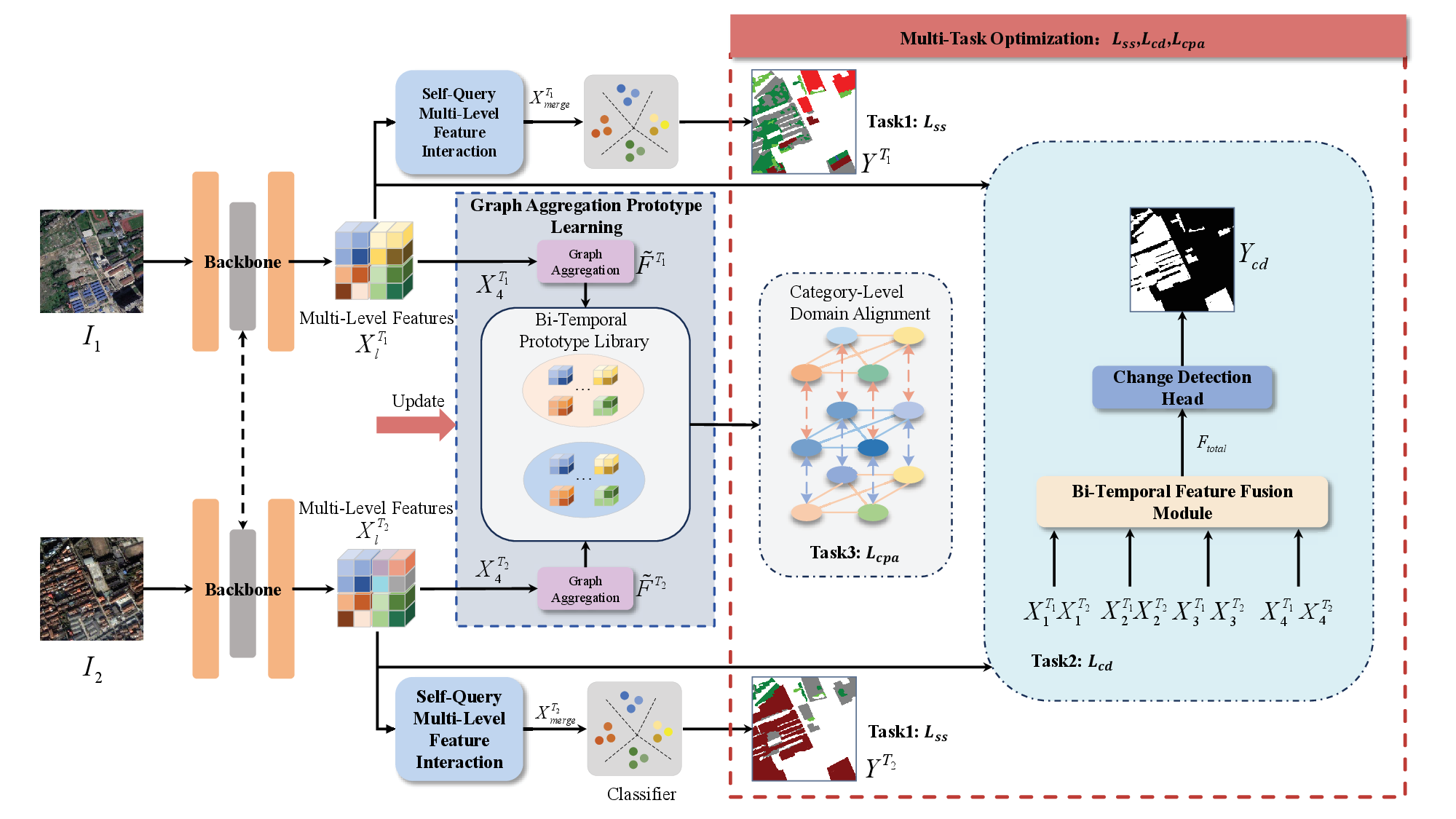}
	\end{tabular}
	\caption{Overall architecture of the proposed Graph-Aggregated Prototype Learning for Remote Sensing Semantic Change Detection (GAPL-SCD).}
	\label{overview}
\end{figure*}
\section{Related Work}
\subsection{Binary Change Detection}
In recent years, many researchers have proposed deep learning-based binary change detection methods, significantly improving detection accuracy \cite{10364762, 9777690, zhang2020deeply, ding2024adapting}. Change detection frameworks based on Convolutional Neural Networks (CNNs) and Transformers have achieved remarkable results in remote sensing image change detection \cite{zhu2024unsupervised, gan2024rfl}. The mainstream CNN-based methods typically rely on encoder-decoder architectures of Siamese networks, combined with attention mechanisms in feature transfer and fusion modules. For example, Chen \emph{et al}. \cite{rs12101662} designed a Siamese network-based spatiotemporal attention neural network, where two branch networks with shared weights extract spatiotemporal features from bi-temporal images. They utilize attention modules in a pyramid structure to capture multi-scale spatiotemporal dependencies, and finally achieve precise change detection results through deep metric learning. Shi \emph{et al}. \cite{9467555} employed a Siamese network to learn the nonlinear transformation of input images, combined with a convolutional attention mechanism to extract more discriminative features, and optimized the distance between changed and unchanged pixels using a contrastive loss function. Li \emph{et al}. \cite{10034814} proposed A2Net, which uses the lightweight MobileNetV2 network to extract deep features and combines a neighborhood aggregation module to fuse features from adjacent stages, thus enhancing temporal feature representation and improving change detection accuracy. 

Transformer-based change detection methods have also made significant progress, particularly when working in conjunction with CNNs. These methods combine local feature extraction and global context modeling to better understand complex changes. Cui \emph{et al}. \cite{9736956} proposed SwinSUNet, which uses a pure Transformer network. The encoder, based on Swin Transformer \cite{liu2021swin} blocks, extracts multi-scale features, and the fusion module combines bi-temporal features. The decoder upsamples the feature map and generates the change detection map. Chen \emph{et al}. \cite{9491802} combined Swin Transformer with CNNs, where CNNs are used to extract high-level semantic features, and the Transformer encoder models the context to generate more contextually informed labels, which are then refined by the decoder to improve the original pixel-level features. Li \emph{et al}. \cite{9761892} introduced TransUNetCD, which combines Transformers with the UNet architecture. The Transformer encoder encodes feature maps extracted by Siamese CNNs, extracting global context information. The decoder uses multi-scale feature fusion and skip connections to restore the change detection map. Additionally, Bandara \emph{et al}. \cite{9883686} proposed a pure Transformer model called ChangeFormer, with its backbone network based on the Siamese SegFormer \cite{xie2021segformer}. In addition, recent studies have also explored the development of novel networks for unsupervised CD \cite{saha2020building} and self-supervised CD \cite{hou2025self}.

\subsection{Semantic Change Detection}
Semantic change detection methods can be mainly divided into three categories according to their image processing workflows: post-classification comparison methods, direct classification methods, and multi-task structural methods. Post-classification comparison methods perform pixel-level classification on multi-temporal remote sensing images and then obtain the results of land cover/land use transitions by comparing classification maps from different dates \cite{6403896}. Although these methods are simple and intuitive, the final accuracy of semantic change detection is affected by the cumulative errors of the classification results, and they do not consider the temporal correlation between multi-temporal images. Direct classification methods transform the semantic change detection task into a multi-classification task based on multi-temporal image data, treating unchanged pixels and each type of change as different categories, thereby directly identifying change regions and classifying the change types. For example, in \cite{varghese2018changenet}, a deep Siamese network was proposed for semantic change detection in street-view images that simultaneously identifies change regions and change types. After fusion and normalization, the model can precisely locate change regions and classify the change type for each pixel. 

Multi-task structural methods not only obtain land cover classification results from multi-temporal remote sensing images but also employ an additional network to extract change masks, which are then combined with the multi-temporal classification results to produce the semantic change detection output. For instance, SSESN \cite{9736642} integrates features from various layers of two encoder branches, with its change detection decoder receiving only the feature differences between the outputs of the two encoder branches as input. Moreover, SSESN introduces feature interaction between the change detection decoder and the semantic segmentation decoder. Daudt \emph{et al}. \cite{daudt2018fully} propose a framework with triple embedding branches, two of the branches segment temporal images into land cover maps, while a binary change detection branch detects the difference information. Yang \emph{et al}. \cite{9555824} propose an asymmetric Siamese network (ASN) to tackle the issue of inconsistent feature extraction in semantic change detection, emphasizing the need for diverse structural modules to capture land-cover variations across temporal images effectively. DEFO-MTLSCD \cite{10422819} is a decoder-focused multi-task learning network that introduces a decoder feature interaction module across sub-tasks. Leveraging the characteristics of a three-branch decoder, it also designs a feature aggregation module that produces more representative shared information by combining the outputs of the last three encoder layers. Chang \emph{et al}. \cite{10466771} simplifies the reasoning process in semantic change detection by jointly utilizing bitemporal features as a unified input, enhancing performance through convolutional attention fusion and refinement modules while avoiding complex difference extraction and branch interaction designs.

\section{Methodology}
\subsection{Overview}
The proposed framework, as illustrated in Fig. \ref{overview}, integrates a multi-task joint optimization approach to simultaneously optimize the primary tasks of semantic segmentation and change detection along with the auxiliary task of graph aggregation prototype learning. This framework comprises five key stages. First, a twin pre-trained convolutional neural network is employed to extract multi-level visual features from bi-temporal images, which are then fed into three distinct branches corresponding to semantic segmentation, change detection, and graph aggregation prototype learning. Next, within the semantic segmentation branch, a self-query multi-level feature interaction module is designed to enhance feature representations via self-query layers, and features are aggregated through hierarchical stacking to bolster the model’s multi-scale perception in complex scenes. Concurrently, a bi-temporal feature fusion module is implemented in the change detection branch; it adopts a composite fusion strategy and utilizes transpose convolution to progressively integrate the fused features from bottom to top, thereby enhancing the model’s ability to identify changes. Furthermore, in the graph aggregation prototype learning branch, an interaction graph is constructed using the final-layer features, and feature interaction and aggregation are performed to obtain class prototypes for each temporal phase. By explicitly constraining the consistency of class prototype relationships across the two time points, category-level domain alignment is achieved, effectively mitigating the impact of irrelevant changes and other confounding factors on the change detection task, which further improves the model’s accuracy and robustness. Finally, a multi-task joint optimization method is proposed that leverages adaptive weight allocation and gradient rotation techniques to alleviate conflicts among training tasks, enabling the joint optimization of all three tasks.

\subsection{Graph Aggregation Prototype Learning}
Assume that ${I_1}\in \mathbb{R}^{H \times W \times 3}$ and ${I_2} \in\mathbb{R}^{H \times W \times 3}$ represent remote sensing images covering the same geographic area at two different time points, where $H$ and $W$ denote the height and width of the images, respectively. In the feature encoding stage, a ResNet-34 pretrained on ImageNet1K\_V1 is employed as the backbone network to extract multi-level features from $I_1$ and $I_2$, denoted as $X_l^{{T_1}} = [X_1^{{T_1}},X_2^{{T_1}},X_3^{{T_1}},X_4^{{T_1}}]$ and $X_l^{{T_2}} = [X_1^{{T_2}},X_2^{{T_2}},X_3^{{T_2}},X_4^{{T_2}}]$, with the embedding dimension of each feature layer being $[64,128,256,512]$. Graph aggregation prototype learning consists of four components: constructing the relational graph, performing graph-based feature interaction and aggregation, obtaining prototypes, and aligning them. The specific process is described below.

First, the final-level features $X_4^{{T_1}} \in \mathbb{R}^{{H_0} \times {W_0} \times d}$ and $X_4^{{T_2}} \in \mathbb{R}^{{H_0} \times {W_0} \times d}$ are flattened into ${F^{{T_1}}},{F^{{T_2}}} \in \mathbb{R}^{{N_s} \times d}$, where ${N_s} = {H_0} \times {W_0}$ is the number of feature vectors, and relational graphs are constructed accordingly. For time point $i \in \left\{ {1,2} \right\}$, a relational graph ${G^{{T_i}}} = \left( {{V^{{T_i}}},{E^{{T_i}}}} \right)$ is defined, where ${V^{{T_i}}}$ represents the set of nodes corresponding to the ${N_s}$ feature vectors in ${F^{{T_i}}}$ and ${E^{{T_i}}} \subseteq {V^{{T_i}}} \times {V^{{T_i}}}$ denotes the set of edges. Here, $d$ represents the embedding dimension of the feature vectors. Since two similar feature vectors are more likely to correspond to the same category, they should be assigned a higher connection weight. To establish this relationship, an adjacency matrix ${A^{{T_i}}} \in \mathbb{R}^{{N_s} \times {N_s}}$ is defined; its computation is based on applying a Gaussian kernel function to the Euclidean distance between two feature vectors:
\begin{equation}
	A_{_{m,n}}^{{T_i}} = \exp \left( { - \frac{{{{\left\| {{f_m} - {f_n}} \right\|}_2}}}{{2{\sigma ^2}}}} \right)
\end{equation}
where ${f_m}$ and ${f_n}$ denote the $m$-th and $n$-th feature vectors in ${F^{{T_i}}}$, $1 \le m,n \le {N_s}$. $\sigma$ denotes the standard deviation parameter controlling the sparsity of ${A^{{T_i}}}$, and ${\left\|  \cdot  \right\|_2}$ denotes the L2 norm.

Then, in the graph-based feature interaction and aggregation stage, due to the sparsity of land cover in remote sensing images, the category instance information represented by a single feature vector is incomplete. To achieve an accurate category feature representation, it is necessary to aggregate the feature vectors that belong to the same category. Utilizing the spatial associations provided by the adjacency matrix, the aggregation of feature embeddings ${F^{{T_i}}}$ is performed using a two-layer graph convolutional network, with the computation formulas for each layer given as follows:
\begin{equation}
	{\tilde F^{{T_i}}} = {\rm{ReLU}}\left( {{{\tilde D}^{ - \frac{1}{2}}}{{\tilde A}^{{T_i}}}{{\tilde D}^{ - \frac{1}{2}}}{F^{{T_i}}}{W^O}} \right)
\end{equation}
\begin{equation}
	{\tilde D_{m,m}} = \sum\limits_{n = 1}^{{N_s}} {\tilde A_{_{m,n}}^{{T_i}}} 
\end{equation}
where ${\tilde A^{{T_i}}} = {A^{{T_i}}} + I$ denotes the graph’s adjacency matrix with self-connections added, the identity matrix $I$ represents the self-connection, $\tilde D \in \mathbb{R}^{{N_s} \times {N_s}}$ denotes the degree matrix of ${\tilde A^{{T_i}}}$, ${W^O}$ is a learnable weight matrix, and ReLU is the non-linear activation function. Through message passing between neighboring feature nodes and iterative feature updating, ${\tilde F^{{T_i}}} \in \mathbb{R}^{{N_s} \times d}$ aggregates node features using the global adjacency matrix of the graph. It enhances each node’s global perception ability, enabling an individual node not only to effectively capture the shared information of similar nodes but also to differentiate heterogeneous nodes in the embedding space, thereby effectively preventing feature confusion.

Finally, to obtain the category prototypes and highlight the node information that is critical for specific classes, the classification confidence for each class is used as a weight, resulting in a prototype that is the weighted average embedding of the feature matrix. 
\begin{equation}
	p_{_k}^{{T_i}} = \frac{{\sum\limits_{m = 1}^{{N_s}} {C_{_{mk}}^{{T_i}} \cdot \tilde F_m^{{T_i}}} }}{{\sum\limits_{m = 1}^{{N_s}} {C_{_{mk}}^{{T_i}}} }}
\end{equation}
where $p_{_k}^{{T_i}} \in \mathbb{R}^d$ denotes the prototype vector for class $k$ at time ${T_i}$, which acts as the proxy for each class during the subsequent domain alignment process between the two-time points. Meanwhile, $C_{_{mk}}^{{T_i}}$ represents the confidence score of the $m$-th feature in ${F^{{T_i}}}$ being classified as class $k$ in the semantic segmentation task. The category prototype library for time $T_i$ is defined as ${P^{{T_i}}} = \left\{ {p_k^{{T_i}}|k = 1,2,...,{N_c}} \right\}$, ${P^{{T_i}}} \in {^{{N_c} \times d}}$, where ${N_c}$ denotes the total number of categories in the remote sensing data. To measure the relationship between any two classes, an affinity matrix is defined, with each element computed as the cosine similarity between the two corresponding category prototypes:
\begin{equation}
	{\cal A}_{i,j}^{{T_1}} = \frac{{P_i^{{T_1}} \cdot P_j^{{T_1}}}}{{{{\left\| {P_i^{{T_1}}} \right\|}_2}{{\left\| {P_j^{{T_1}}} \right\|}_2}}}
\end{equation}
\begin{equation}
	{\cal A}_{i,j}^{{T_2}} = \frac{{P_i^{{T_2}} \cdot P_j^{{T_2}}}}{{{{\left\| {P_i^{{T_2}}} \right\|}_2}{{\left\| {P_j^{{T_2}}} \right\|}_2}}}
\end{equation}
\begin{equation}
	{\cal A}_{i,j}^{{T_{1,2}}} = \frac{{P_i^{{T_1}} \cdot P_j^{{T_2}}}}{{{{\left\| {P_i^{{T_1}}} \right\|}_2}{{\left\| {P_j^{{T_2}}} \right\|}_2}}}
\end{equation}
where ${\cal A}^{{T_1}} \in \mathbb{R}^{{N_c} \times {N_c}}$, ${\cal A}^{{T_2}} \in \mathbb{R}^{{N_c} \times {N_c}}$, and ${\cal A}^{{T_{1,2}}} \in \mathbb{R}^{{N_c} \times {N_c}}$ denote the prototype relationship affinity matrices for time $T_1$, time $T_2$, and across temporal phases, respectively, $1 \le i,j \le {N_c}$. Since the intrinsic relationships between classes should remain invariant across different time points during the feature extraction process, using prototypes as proxies for classes and maintaining the consistency of prototype relationships between the two time points can explicitly transfer the knowledge of intra-class and inter-class relationships from time $T_1$ to time $T_2$. Specifically, a loss function ${L_1}$ is adopted to measure the differences between ${\cal A}^{{T_1}}$ and ${\cal A}^{{T_2}}$, between ${\cal A}^{{T_1}}$ and ${\cal A}^{{T_1,2}}$, and between ${\cal A}^{{T_2}}$ and ${\cal A}^{{T_1,2}}$, thereby constraining the consistency of the prototype relationships to achieve domain alignment. Therefore, the objective function for the graph aggregation prototype learning task is defined as follows:
\begin{equation}
	\begin{array}{l}
		{L_{cpa}} = {L_1}\left( {{\cal A}_{}^{{T_1}} - {\cal A}_{}^{{T_2}}} \right) + {L_1}\left( {{\cal A}_{}^{{T_1}} - {\cal A}_{}^{{T_{1,2}}}} \right) + {L_1}\left( {{\cal A}_{}^{{T_2}} - {\cal A}_{}^{{T_{1,2}}}} \right)\\
		{\rm{      }} = \frac{1}{{{N_c} \times {N_c}}}\sum\limits_{i = 1}^{{N_c}} {\sum\limits_{j = 1}^{{N_c}} {\left( {\left| {{\cal A}_{i,j}^{{T_1}} - {\cal A}_{i,j}^{{T_2}}} \right| + \left| {{\cal A}_{i,j}^{{T_1}} - {\cal A}_{i,j}^{{T_{1,2}}}} \right| + \left| {{\cal A}_{i,j}^{{T_2}} - {\cal A}_{i,j}^{{T_{1,2}}}} \right|} \right)} } 
	\end{array}
\end{equation}
In the above objective function, the affinity matrix is computed using the global prototype library maintained during model training, denoted as $P_{^{global}}^{{T_i}}$. Moreover, during training, a local prototype library $P_{^{local}}^{{T_i}}$ is generated for each mini-batch, and the global prototype should remain consistent with the local prototypes. Thus, the local prototypes are used to update the global prototype. The update formula is given by:
\begin{equation}
	P_{^{global}}^{{T_i}} = \beta P_{^{global}}^{{T_i}} + \left( {1 - \beta } \right)P_{^{local}}^{{T_i}}
\end{equation}
where $\beta$ is the momentum coefficient used to control the smoothness of the update, with $\beta = 0.9$ in this paper.

\subsection{Self-Query Multi-Level Feature Interaction}
\begin{figure}[htbp]
	\centering
	\begin{tabular}{@{}c@{}c}
		\includegraphics[width=1\linewidth]{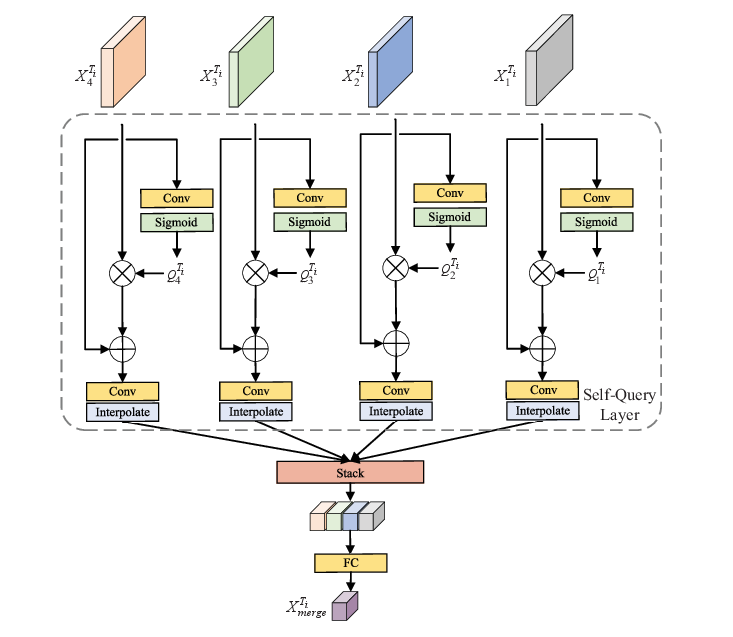}
	\end{tabular}
	\caption{Self-Query Multi-Level Feature Interaction Module (SQMLFI).}
	\label{SQMLFI}
\end{figure}

In the semantic segmentation branch, to enhance the model’s multi-scale perception in complex scenes, multi-level features from the backbone network are exploited, enhanced with self-query layers, and finally fused via hierarchical stacking, as shown in Fig. \ref{SQMLFI}. For the $l$-th layer feature at time $T_i$, a self-attention-like mechanism is applied to enhance the feature, which encourages the network to focus on the regions of interest. Finally, features from different levels are resized via interpolation to a unified spatial dimension and then stacked for fusion output. This module enables each level not only to learn the feature information at its inherent scale more effectively but also to obtain more diverse representations through the fusion of multiple scales.

Specifically, the multi-level features $X_l^{{T_i}}=[X_1^{{T_i}},X_2^{{T_i}},X_3^{{T_i}},X_4^{{T_i}}]$ from time $T_i$ are separately fed into convolutional layers and passed through a Sigmoid activation function to obtain the self-query attention scores:
\begin{equation}
	Q_l^{{T_i}} = {\rm{Sigmoid}}\left( {{\rm{Conv}}\left( {X_l^{{T_i}}} \right)} \right)
\end{equation}
where the convolutional layer Conv uses a kernel of $3 \times 3$, with a padding of 1 and a stride of 1. The obtained attention scores are then used to weight the original features, enhancing the network’s focus on the regions of interest. Moreover, to stabilize training, a residual connection mechanism is introduced. The self-query enhanced features are then passed through another convolutional layer for further information refinement. Subsequently, using bilinear interpolation, they are resized to match the spatial dimensions of a reference feature map $X_1^{{T_i}}$, which facilitates subsequent stacking:
\begin{equation}
	\mathord{\buildrel{\lower3pt\hbox{$\scriptscriptstyle\frown$}} 
		\over X}{ _l^{T_i}} = {\rm{Interpolate}}\left( {{\rm{Conv}}\left( {X_l^{{T_i}} \odot Q_l^{{T_i}} + X_l^{{T_i}}} \right)} \right)
\end{equation}
where $\mathord{\buildrel{\lower3pt\hbox{$\scriptscriptstyle\frown$}} \over X} {_l^{T_i}} \in \mathbb{R}^{\frac{H}{4} \times \frac{W}{4} \times 256}$, $\odot $ denotes the Hadamard (element-wise) product, Interpolate denotes bilinear interpolation, and the convolutional block Conv comprises a convolutional layer with a kernel of $3 \times 3$ and padding set to 1, followed by a ReLU activation and a BatchNorm layer.

Finally, to fuse the multi-level features, the four features are stacked along a new dimension and passed through a fully connected (FC) layer to facilitate multi-level feature interaction. This operation assigns fusion weights to each level to extract more representative features:
\begin{equation}
	X_{merge}^{{T_i}} = {\rm{FC}}\left( {{\rm{stack}}\left( {\mathord{\buildrel{\lower3pt\hbox{$\scriptscriptstyle\frown$}} 
				\over X}{_1^{T_i}},\mathord{\buildrel{\lower3pt\hbox{$\scriptscriptstyle\frown$}} 
				\over X} {_2^{T_i}},\mathord{\buildrel{\lower3pt\hbox{$\scriptscriptstyle\frown$}} 
				\over X} {_3^{T_i}},\mathord{\buildrel{\lower3pt\hbox{$\scriptscriptstyle\frown$}} 
				\over X}{_4^{T_i}}} \right)} \right)
\end{equation}
where $X_{merge}^{{T_i}} \in \mathbb{R}^{\frac{H}{4} \times \frac{W}{4} \times 256}$, stack represents the stacking operation, and FC denotes the fully connected layer.
\subsection{Bi-Temporal Feature Fusion Module}
\begin{figure}[htbp]
	\centering
	\begin{tabular}{@{}c@{}c}
		\includegraphics[width=0.9\linewidth]{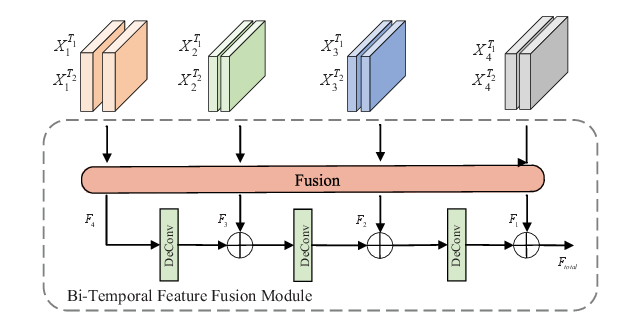}
	\end{tabular}
	\caption{Bi-Temporal Feature Fusion Module (BTFF).}
	\label{BTFF}
\end{figure}

In the change detection branch, a bi-temporal feature fusion module is designed that adopts a composite fusion strategy. By employing transpose convolution, the fused features are progressively integrated from bottom to top, thereby enhancing the model’s ability to recognize changes. As shown in Fig. \ref{BTFF}, specifically for the fusion of bi-temporal feature pairs $X_l^{{T_1}},X_l^{{T_2}}\left( {s = 1,2,3,4} \right)$ across four scales, a composite fusion method is used that measures the differences between the two temporal images through both difference values and cosine similarity, thereby extracting a more robust representation of change information:
\begin{equation}
	D = {\rm{concat}}\left( {\left[ {{\rm{Con}}{{\rm{v}}_{3 \times 3}}\left( {X_l^{{T_2}} - X_l^{{T_1}}} \right),{\rm{Con}}{{\rm{v}}_{3 \times 3}}\left( {\cos \left( {X_l^{{T_2}},X_l^{{T_1}}} \right)} \right)} \right]} \right)
\end{equation}
\begin{equation}
	{F_l} = {\rm{Con}}{{\rm{v}}_{1 \times 1}}\left( {{\rm{ReLU}}\left( {{\rm{BN}}\left( {{\rm{Con}}{{\rm{v}}_{3 \times 3}}\left( D \right)} \right)} \right)} \right)
\end{equation}
where $D$ denotes the difference information extracted after fusion, $\cos \left(  \cdot  \right)$ denotes the computation of cosine similarity, and $F_l$ denotes the refined information obtained after the fusion of the $l$-th level features.

To integrate the fused information from the four scales, transpose convolution is employed for upsampling. Following the approach of a feature pyramid, $F_l$ is progressively integrated from bottom to top to improve change detection accuracy:
\begin{equation}
	{F_{total}} = {\rm{DeConv}}\left( {{\rm{DeConv}}\left( {{\rm{DeConv}}\left( {{F_4}} \right) + {F_3}} \right) + {F_2}} \right) + {F_1}
\end{equation}
where DeConv denotes the transpose convolution.
\subsection{Multi-Task Optimization}
For the semantic segmentation task, the outputs of the self-query multi-level feature interaction module, denoted as $X_{merge}^{{T_1}}$ and $X_{merge}^{{T_2}}$, are obtained. These features are then passed through two convolutional layers with a kernel size of $1 \times 1$ for classification, yielding the ground cover prediction probabilities for the bi-temporal remote sensing images, denoted as ${\tilde Y^{{T_1}}}$ and ${\tilde Y^{{T_2}}}$. Assuming that the ground truth maps for ground cover in the bi-temporal images are ${Y^{{T_1}}}$ and ${Y^{{T_2}}}$, a cross-entropy loss function is used to quantify the difference between the predictions and the ground truth. Specifically, the loss function is mathematically expressed as follows:
\begin{equation}
	\begin{array}{l}
		{L_{ss}} = {{\left( {L_{_{ss}}^{{T_1}} + L_{_{ss}}^{{T_2}}} \right)} \mathord{\left/
				{\vphantom {{\left( {L_{_{ss}}^{{T_1}} + L_{_{ss}}^{{T_2}}} \right)} 2}} \right.
				\kern-\nulldelimiterspace} 2}\\
		{\rm{    }} = {{\left( { - \frac{1}{{H \times W}}\sum\limits_{j = 1}^{H \times W} {\sum\limits_{c = 1}^{{N_c}} {y_{_{i,c}}^{{T_1}}\log \left( {\tilde y_{_{j,c}}^{{T_1}}} \right)} }  - \frac{1}{{H \times W}}\sum\limits_{j = 1}^{H \times W} {\sum\limits_{c = 1}^{{N_c}} {y_{_{j,c}}^{{T_2}}\log \left( {\tilde y_{_{j,c}}^{{T_2}}} \right)} } } \right)} \mathord{\left/
				{\vphantom {{\left( { - \frac{1}{{H \times W}}\sum\limits_{j = 1}^{H \times W} {\sum\limits_{c = 1}^{{N_c}} {y_{_{i,c}}^{{T_1}}\log \left( {\tilde y_{_{j,c}}^{{T_1}}} \right)} }  - \frac{1}{{H \times W}}\sum\limits_{j = 1}^{H \times W} {\sum\limits_{c = 1}^{{N_c}} {y_{_{j,c}}^{{T_2}}\log \left( {\tilde y_{_{j,c}}^{{T_2}}} \right)} } } \right)} 2}} \right.
				\kern-\nulldelimiterspace} 2}
	\end{array}
\end{equation}
where $y_{j,c}^{{T_i}}$ indicates whether the $j$-th pixel at time $T_i$ truly belongs to class $c$, and $\tilde y_{j,c}^{{T_i}}$ represents the probability predicted by the model that pixel $j$ belongs to class $c$.

For the change detection task, the output from the bi-temporal feature fusion module, denoted as ${F_{total}}$, is obtained. By passing ${F_{total}}$ through two convolutional layers with kernel sizes of $3 \times 3$ and $1 \times 1$, respectively, the model produces the predicted change mask for the bi-temporal remote sensing images, denoted as ${\tilde Y_{cd}}$. Assuming that the ground truth for changes is ${Y_{cd}}$, cross-entropy loss is similarly applied to quantify the difference:
\begin{equation}
	{L_{cd}} =  - \frac{1}{{H \times W}}\sum\limits_{j = 1}^{H \times W} {\sum\limits_{c = 1}^2 {y_{_{i,c}}\log \left( {\tilde y_{_{j,c}}} \right)} } 
\end{equation}
where since change detection is a binary classification problem, the number of classes is 2.

In this paper, the primary tasks—semantic segmentation and change detection—and the auxiliary task, graph aggregation prototype learning, share the parameters of the backbone network. When deciding how to update the parameters in the shared backbone, multiple objectives must be considered. Since the three tasks have different output dimensions and types of training loss functions, it is difficult to guarantee that all tasks are well aligned during training; as a result, the shared parameters may become biased toward one task. If the gradient magnitude of one task is much larger than that of another, it will dominate the averaged gradient, causing the backbone’s parameter updates to improve the performance of the dominant task while degrading the performance of the subordinate one—an effect known as negative transfer.

To address this, a multi-task joint optimization method is designed. Based on the characteristics of each task, adaptive weight allocation and a gradient rotation method are employed to alleviate conflicts among training tasks, thereby achieving joint optimization training for all three tasks.

Specifically, for the semantic segmentation and change detection tasks, which are both essentially classification tasks, their loss functions are relatively similar and can be merged. The merging method uses homoscedastic uncertainty to adaptively allocate weights to the losses of these two tasks:
\begin{equation}
	{L_{merge}} = \frac{1}{{2\sigma _{^1}^2}}{L_{ss}} + \frac{1}{{2\sigma _{^2}^2}}{L_{cd}} + \ln \left( {1 + \sigma _{^1}^2} \right) + \ln \left( {1 + \sigma _{^2}^2} \right)
\end{equation}
Here, ${\sigma _1}$ and ${\sigma _2}$ denote the observation noise and serve as learnable parameters for the weights of the two task losses, $\ln \left( {1 + \sigma _{^1}^2} \right)$ and $\ln \left( {1 + \sigma _{^2}^2} \right)$ are regularization terms added to prevent training collapse. This design provides a principled approach to learning the relative weights of the various loss functions; since each loss function is smooth and differentiable, it prevents the task weights from converging to zero. In contrast, if one were to directly use a simple linear sum of the losses to learn the weights, the weights might rapidly converge to zero.

After adaptive weight allocation, the losses for the semantic segmentation and change detection tasks are unified into a single classification loss, denoted as ${L_{merge}}$. In contrast, the graph aggregation prototype learning task is a regression task whose loss function type differs significantly from that of the classification tasks; therefore, adaptive weight allocation is not applied for its optimization.

To further alleviate conflicts between the gradients associated with ${L_{merge}}$ and ${L_{cpa}}$, this paper introduces a gradient rotation method. The idea is defined as follows: if the directions of two gradients are opposite—that is, if their cosine similarity is negative—then these two gradients are considered to be in conflict. When a conflict occurs between the gradients of two tasks, the gradient rotation method rotates each task’s gradient onto the normal plane of the other task’s gradient. This process effectively removes the conflicting component, ensuring that each task’s gradient has zero contribution in the conflicting direction, so that each task only updates the model weights along directions that do not conflict with the gradients of the other tasks.

The implementation is as follows: First, assume that the gradient of the task represented by ${L_{merge}}$ is a vector ${g_a} = {\nabla _\theta }{L_{merge}}\left( \theta  \right)$ and the gradient of the task represented by ${L_{cpa}}$ is a vector ${g_b} = {\nabla _\theta }{L_{cpa}}\left( \theta  \right)$. Then, for $g_a$, the cosine similarity between $g_a$ and $g_b$ is computed. If the cosine similarity is negative, $g_a$ is replaced by its projection onto the normal plane of ${g_b}$.
\begin{equation}
	g'_a =
	\begin{cases}
		g_a - \frac{g_a \cdot g_b}{\|g_b\|_2^2}\,g_b, & \text{if } \frac{g_a \cdot g_b}{\|g_a\|_2\,\|g_b\|_2} < 0, \\[1ex]
		g_a, & \text{otherwise.}
	\end{cases}
\end{equation}

If the cosine similarity value is non-negative, $g_a$ remains unchanged. For $g_b$, the same process is repeated. The corrected gradients are ${g'_a}$ and ${g'_b}$; the gradients of the two tasks become orthogonal to the conflicting direction of the other task's gradient. Therefore, the final objective function and optimization method for multi-task optimization in this section can be expressed as follows:
\begin{equation}
	{L_{total}} = {L_{merge}} + {L_{cpa}}
\end{equation}
\begin{equation}
	\Delta \theta  =  - {\rm{lr}} \cdot {\rm{project}}\left( {{g_a},{g_b}} \right) =  - {\rm{lr}} \cdot \left( {{g'_a} + {g'_b}} \right)
\end{equation}
where project denotes the optional gradient rotation operation, $\Delta \theta$ represents the parameter $\theta$ update in one iteration, lr denotes the learning rate. After computing the gradients for the two loss components separately, they are corrected using the modified gradients, and the parameters are updated with these corrected gradients. The gradient rotation method ensures that when the gradients do not conflict, they remain unchanged; however, when conflicts occur, each task’s gradient is modified to minimize the negative interference from the gradients of the other tasks.

\section{Experiments}
In this section, we first introduce the experimental parameter settings used during training and provide detailed information about the datasets. To validate the effectiveness of the proposed multi-task remote sensing image semantic change detection method guided by graph aggregation prototype learning, experiments were conducted on two publicly available remote sensing image semantic change detection datasets—SECOND \cite{9555824} and Landsat-SCD \cite{Yuan2022ATS}—and compared with other state-of-the-art methods.
\subsection{Datasets}
\begin{figure}[htbp]
	\centering
	\begin{tabular}{@{}c@{}c}
		\includegraphics[width=1\linewidth]{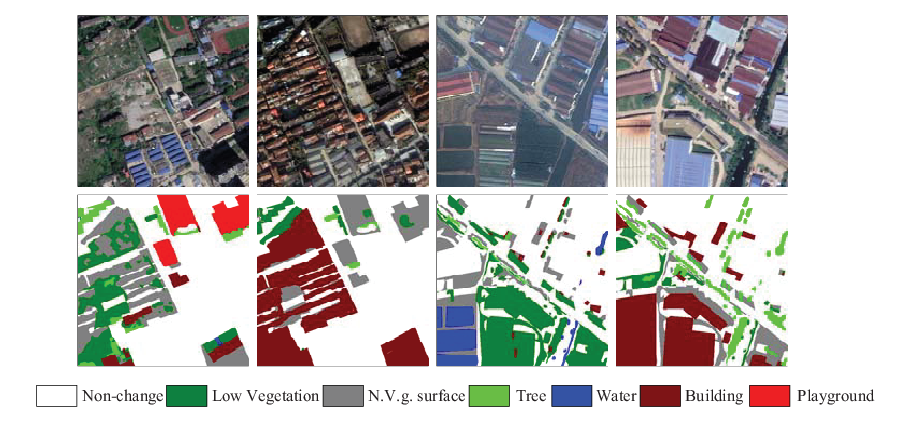}
	\end{tabular}
	\caption{A partial image visualization of the SECOND Dataset.}
	\label{SECOND-Dataset}
\end{figure}

1) \textit{SECOND Dataset:} This dataset consists of 4,662 pairs of multi-temporal high-resolution remote sensing images covering areas such as Shanghai, Chengdu, and Hangzhou. The spatial resolution of each image pair ranges from 0.5 to 3 m, and each image has a size of $512 \times 512$ pixels. The land cover classification maps in this dataset include one “no change” category and six land cover categories: water bodies, non-vegetated surfaces, low vegetation, trees, buildings, and sports fields. In the public version of the dataset, the training set comprises 2,968 image pairs, and the test set comprises 1,694 image pairs. Fig. \ref{SECOND-Dataset} shows a visualization of some sample images from the SECOND dataset.

\begin{figure}[htbp]
	\centering
	\begin{tabular}{@{}c@{}c}
		\includegraphics[width=0.9\linewidth]{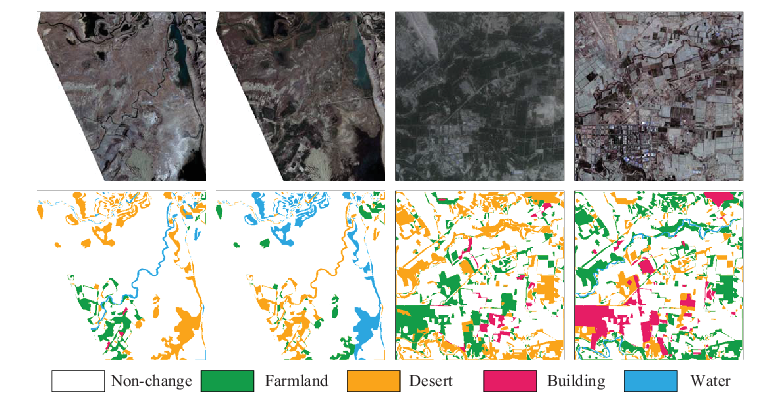}
	\end{tabular}
	\caption{A partial image visualization of the Landsat-SCD Dataset.}
	\label{Landsat-SCD-Dataset}
\end{figure}

2) \textit{Landsat-SCD Dataset:} This dataset is constructed from Landsat satellite images collected between 1990 and 2020. The observation area is located in Tumushuke, Xinjiang, China, on the edge of the Taklamakan Desert. Each image pair in this dataset has a spatial resolution of 30 m. The dataset is annotated with a “no change” category and four land cover categories, including farmland, desert, buildings, and water bodies (only the regions with changes are annotated). The Landsat-SCD dataset originally consists of 8,468 image pairs, and each image has a size of $416 \times 416$ pixels. After removing redundant samples obtained through data augmentation in the public dataset, 2,425 original image pairs remain. For validating the proposed method, the training set contains 1,455 image pairs, the validation set contains 485 image pairs, and the test set contains 485 image pairs. Fig. \ref{Landsat-SCD-Dataset} presents a visualization of some sample images from the Landsat-SCD dataset.

\subsection{Experimental Setup}
1) \textit{Assessment Criteria:} Four metrics commonly used in semantic change detection tasks are employed to quantitatively evaluate segmentation accuracy: Overall Accuracy (OA), mean Intersection over Union (mIoU), and two metrics specifically designed for semantic change detection—the Separation Kappa coefficient (SeK) and the F1 score (${F_{scd}}$). Let $M = \left\{ {{m_{ij}}} \right\}$ be the confusion matrix, where ${m_{ij}}$ represents the number of pixels predicted as class $i$ while the true class is $j$, with class 0 representing the “no change” category. In semantic change detection tasks, due to the dominance of the “no change” class, OA cannot effectively evaluate the model's ability to distinguish between semantic categories. Therefore, mIoU and SeK are introduced to respectively assess the performance of change detection and semantic segmentation. 

2) \textit{Comparative Analysis with State-of-the-Art Algorithms:} Several typical remote sensing image semantic change detection methods were used as baselines: SSESN \cite{9736642}, SCDNet \cite{PENG2021102465}, SSCD-l \cite{9721305}, Bi-SRNet \cite{9721305}, DEFO-MTLSCD \cite{10422819}, TED \cite{10443352}, and SCanNet \cite{10443352}. SSESN interprets change data by integrating a pyramid structure, then uses this integration to assign position priority to the two temporal branches. SCDNet employs a dual-branch encoder–decoder architecture that inputs bi-temporal images into the encoder to obtain multi-scale deep representations. A multi-scale convolution unit is used to expand the receptive field and capture multi-scale information; the generated feature maps are merged with the encoder features and fed into the decoder. Bi-SRNet leverages self-attention to support deep semantic information and then implements a cross-temporal module to facilitate interaction between the two temporal branches. As the base model of Bi-SRNet, SSCD-l significantly enhances the semantic detection capability of the dual branches through a semantic consistency loss. DEFO-MTLSCD is a decoder-focused multi-task learning network that introduces a decoder feature interaction module across subtasks, utilizes a three-branch decoder architecture, and designs a feature aggregation module to produce more representative shared information by merging the outputs of the last three encoder layers. SCanNet proposes a semantic change transformer to model the semantic transformation between bi-temporal remote sensing images, and then introduces a semantic learning scheme that employs spatiotemporal constraints consistent with the semantic change detection task to guide the learning of semantic changes. TED, as the base model of SCanNet, extracts temporal features and change representations through a three-branch encoder–decoder network.

3) \textit{Implementation Details:} All comparative experiments for the proposed multi-task remote sensing image semantic change detection method guided by graph aggregation prototype learning were implemented using the PyTorch 2.2.0 framework. These experiments were conducted on an NVIDIA RTX 4090 GPU. During training, the learning rate was set to 0.0001, the batch size to 8, and the model was trained for 50 epochs. A cosine annealing function was used to dynamically adjust the learning rate throughout the training process. The Adam optimizer was employed to optimize all network parameters with a weight decay coefficient of 1e-6. For the comparative methods, the training parameters were set according to the experimental configurations described in their respective papers.

\subsection{Quantitative and Qualitative Experimental Results}
Table \ref{result-second} presents the experimental accuracies of the proposed method and the various comparison methods on the SECOND dataset. Table \ref{result-landsat-scd} shows the experimental accuracies of the proposed method and the comparison methods on the Landsat-SCD dataset. In both tables, the best classification performance is indicated using bold text.
\begin{table*}[htbp]
	\renewcommand{\arraystretch}{1.3}
	\centering
	\caption{Comparison of different methods on the SECOND dataset.}
	\label{result-second}
	\resizebox{\textwidth}{!}{
		\begin{tabular}{c|ccccccc|c}
			\hline
			\textbf{Category} 
			& \textbf{SSESN \cite{9736642}} 
			& \textbf{SCDNet \cite{PENG2021102465}} 
			& \textbf{SSCD-l \cite{9721305}} 
			& \textbf{Bi-SRNet \cite{9721305}} 
			& \textbf{DEFO-MTLSCD \cite{10422819}} 
			& \textbf{TED \cite{10443352}} 
			& \textbf{SCanNet \cite{10443352}} 
			& \textbf{Ours} \\
			\hline
			
			Non-change     
			& 91.33 & 92.22 & 92.88 & 92.58 & 91.55 & 92.00 & \textbf{93.28} & 92.95 \\
			
			Low vegetation 
			& 50.88 & 52.51 & 49.89 & 55.25 & 49.03 & 55.29 & 52.01 & \textbf{56.42} \\
			
			N.v.g. surface 
			& 56.63 & 56.78 & 59.33 & 62.31 & 57.11 & 61.76 & 60.16 & \textbf{63.03} \\
			
			Tree            
			& 35.65 & 45.60 & 42.75 & 43.44 & 38.99 & \textbf{46.18} & 41.11 & 43.28 \\
			
			Water           
			& 45.52 & 56.20 & 58.73 & \textbf{66.78} & 45.28 & 53.98 & 60.12 & 57.57 \\
			
			Building        
			& 66.26 & 68.32 & 70.26 & 73.70 & 71.75 & 71.02 & 74.42 & \textbf{74.62} \\
			
			Playground      
			& 43.31 & \textbf{73.45} & 71.26 & 69.58 & 62.35 & 70.30 & 70.68 & 70.60 \\
			
			\hline
			
			OA (\%)         
			& 85.29 & 86.02 & 86.81 & 87.26 & 85.84 & 86.91 & 87.05 & \textbf{87.56} \\
			
			mIoU (\%)       
			& 68.56 & 70.48 & 72.21 & 72.34 & 69.75 & 71.30 & 72.78 & \textbf{73.03} \\
			
			SeK (\%)        
			& 14.78 & 17.51 & 20.94 & 21.17 & 15.92 & 19.14 & 21.92 & \textbf{22.62} \\
			
			${F_{scd}}$ (\%) 
			& 55.34 & 58.04 & 61.12 & 61.44 & 56.10 & 59.83 & 61.80 & \textbf{62.75} \\
			
			\hline
		\end{tabular}
	}
\end{table*}

From the data in Table \ref{result-second}, it can be observed that the proposed method achieves the best results across multiple categories and evaluation metrics, demonstrating its exceptional ability to handle fine-grained variations in complex scenarios. Specifically, compared with existing methods, the proposed approach exhibits significant improvements in categories such as "low vegetation," "n.v.g surface," and "buildings," highlighting its robustness in handling different land cover types. For the "low vegetation" category, the proposed method achieves an accuracy of 56.42\%, outperforming Bi-SRNet and TED, with improvements of 5.54\% and 3.91\% compared to SSESN and SCDNet, respectively. In the "n.v.g surface" category, the proposed method attains an accuracy of 63.03\%, surpassing all other methods, with an increase of 6.25\% over SCDNet and 0.72\% over Bi-SRNet. For the "buildings" category, the proposed method achieves an accuracy of 74.62\%, outperforming all competing methods, slightly surpassing the second-best model, SCanNet, further proving its superior performance in urban change detection.
Compared with TED, OA is improved by 0.65\%, and compared with SCanNet, it is improved by 0.51\%, demonstrating the superior recognition capability of the proposed method in comprehensive change detection tasks. In terms of mean Intersection over Union (mIoU), the proposed method achieves 73.03\%, surpassing other methods by 0.25\% over SCanNet and 0.69\% over Bi-SRNet, indicating its outstanding performance across all change detection categories.
For semantic category consistency (SeK), the proposed method achieves 22.62\%, improving by 0.70\% over SCanNet and 3.48\% over TED, demonstrating its stronger capability in semantic category recognition. Regarding ${F_{scd}}$, the proposed method attains the highest value of 62.75\%, surpassing SCanNet by 0.95\% and SSCD-l by 1.63\%, achieving the best performance in the semantic segmentation accuracy of change regions.

\begin{table*}[htbp]
	\renewcommand{\arraystretch}{1.3}
	\centering
	\caption{Comparison of different methods on the Landsat-SCD dataset.}
	\label{result-landsat-scd}
	\resizebox{\textwidth}{!}{
	\begin{tabular}{c|ccccccc|c}
		\hline
		\textbf{Category} & \textbf{SSESN \cite{9736642}} & \textbf{SCDNet \cite{PENG2021102465}} & \textbf{SSCD-l \cite{9721305}} & \textbf{Bi-SRNet \cite{9721305}} & \textbf{DEFO-MTLSCD \cite{10422819}} & \textbf{TED \cite{10443352}} & \textbf{SCanNet \cite{10443352}} & \textbf{Ours} \\
		\hline
		Non-change   & 95.10 & 95.61 & 96.90 & 96.75 & 97.39 & 97.41 & 97.84 & \textbf{97.63} \\
		Farmland    & 63.07 & 71.24 & 75.72 & 79.26 & 80.32 & 79.97 & 81.91 & \textbf{83.65} \\
		Desert      & 66.74 & 74.76 & 78.76 & 81.31 & 82.71 & 82.57 & 83.91 & \textbf{85.75} \\
		Building    & 40.18 & 59.45 & 71.22 & 77.22 & 70.32 & 74.47 & 77.62 & \textbf{81.21} \\
		Water       & 85.49 & 84.96 & 86.98 & 88.23 & 88.85 & 87.29 & 89.13 & \textbf{89.99} \\
		\hline
		OA (\%)         & 89.15 & 91.44 & 93.20 & 93.80 & 94.41 & 94.39 & 95.04 & \textbf{95.30} \\
		mIoU (\%)       & 74.17 & 77.95 & 81.89 & 82.94 & 85.07 & 84.79 & 86.37 & \textbf{87.02} \\
		SeK (\%)        & 24.28 & 32.46 & 41.77 & 44.27 & 48.40 & 48.33 & 52.63 & \textbf{53.88} \\
		${F_{scd}}$ (\%) & 68.27 & 74.82 & 80.53 & 82.01 & 83.34 & 83.63 & 85.62 & \textbf{85.99} \\
		\hline
	\end{tabular}}
\end{table*}

From the data in Table \ref{result-landsat-scd}, it can be observed that the proposed method achieves significant improvements across multiple categories and evaluation metrics, demonstrating its strong capability in handling semantic change detection tasks for remote sensing images.
For the "farmland" category, the proposed method achieves an accuracy of 83.65\%, significantly outperforming the second-best method, SCanNet, with an improvement of 4.39\% over Bi-SRNet and 3.33\% over DEFO-MTLSCD. In the "desert" category, the proposed method attains an accuracy of 85.75\%, surpassing SCanNet by 1.84\% and also outperforming DEFO-MTLSCD and TED, indicating superior detection ability in complex textured surfaces.
For the "building" category, the proposed method achieves 81.21\% accuracy, substantially leading all other methods, with an improvement of 3.99\% over Bi-SRNet and 6.74\% over TED, demonstrating its effectiveness in complex structural regions. Additionally, in the "water body" category, the proposed method attains an accuracy of 89.99\%, surpassing SCanNet by 0.86\%, showing its higher accuracy in detecting highly reflective regions such as water bodies.

\begin{figure*}[htbp]
	\centering
	\begin{tabular}{@{}c@{}c}
		\includegraphics[width=1\linewidth]{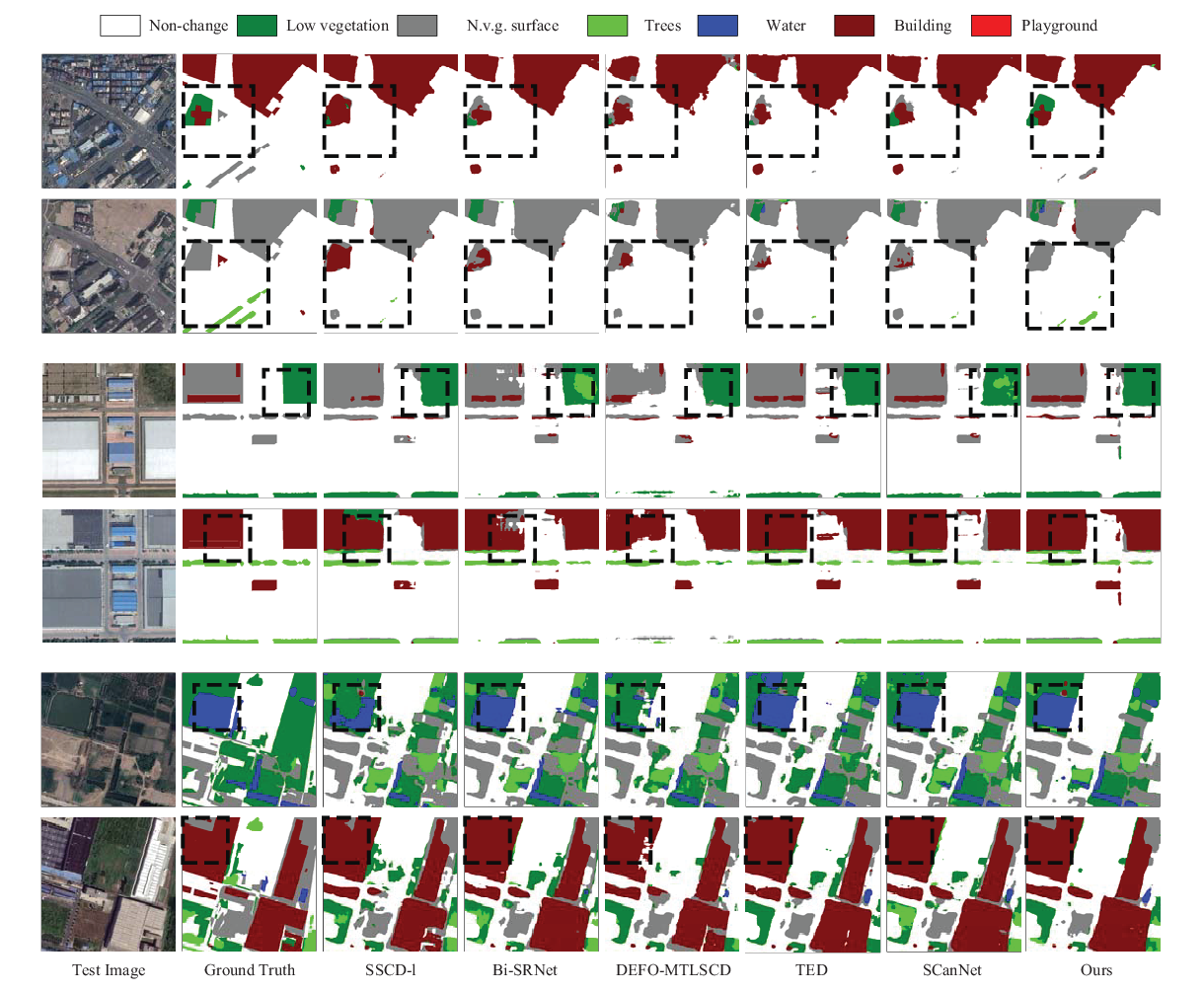}
	\end{tabular}
	\caption{Qualitative experimental results on the SECOND dataset.}
	\label{visual-second}
\end{figure*}

Regarding overall evaluation metrics, the proposed method achieves an OA of 95.30\%, surpassing the second-best SCanNet by 0.26\% and Bi-SRNet by 1.50\%, indicating its high overall detection accuracy across different change detection categories. In terms of mIoU, the proposed method achieves 87.02\%, outperforming all other methods, with an improvement of 0.65\% over SCanNet and 1.95\% over DEFO-MTLSCD, demonstrating its precision and robustness in segmenting different change regions.
For semantic category consistency (SeK), the proposed method reaches 53.88\%, significantly surpassing the second-best SCanNet and improving by 5.48\% over DEFO-MTLSCD, highlighting its advantages in reducing false alarms and enhancing change detection effectiveness. Regarding ${F_{scd}}$, the proposed method achieves 85.99\%, slightly outperforming the second-best SCanNet, further validating its superior accuracy in change region detection.

\begin{figure*}
	\centering
	\begin{tabular}{@{}c@{}c}
		\includegraphics[width=1\linewidth]{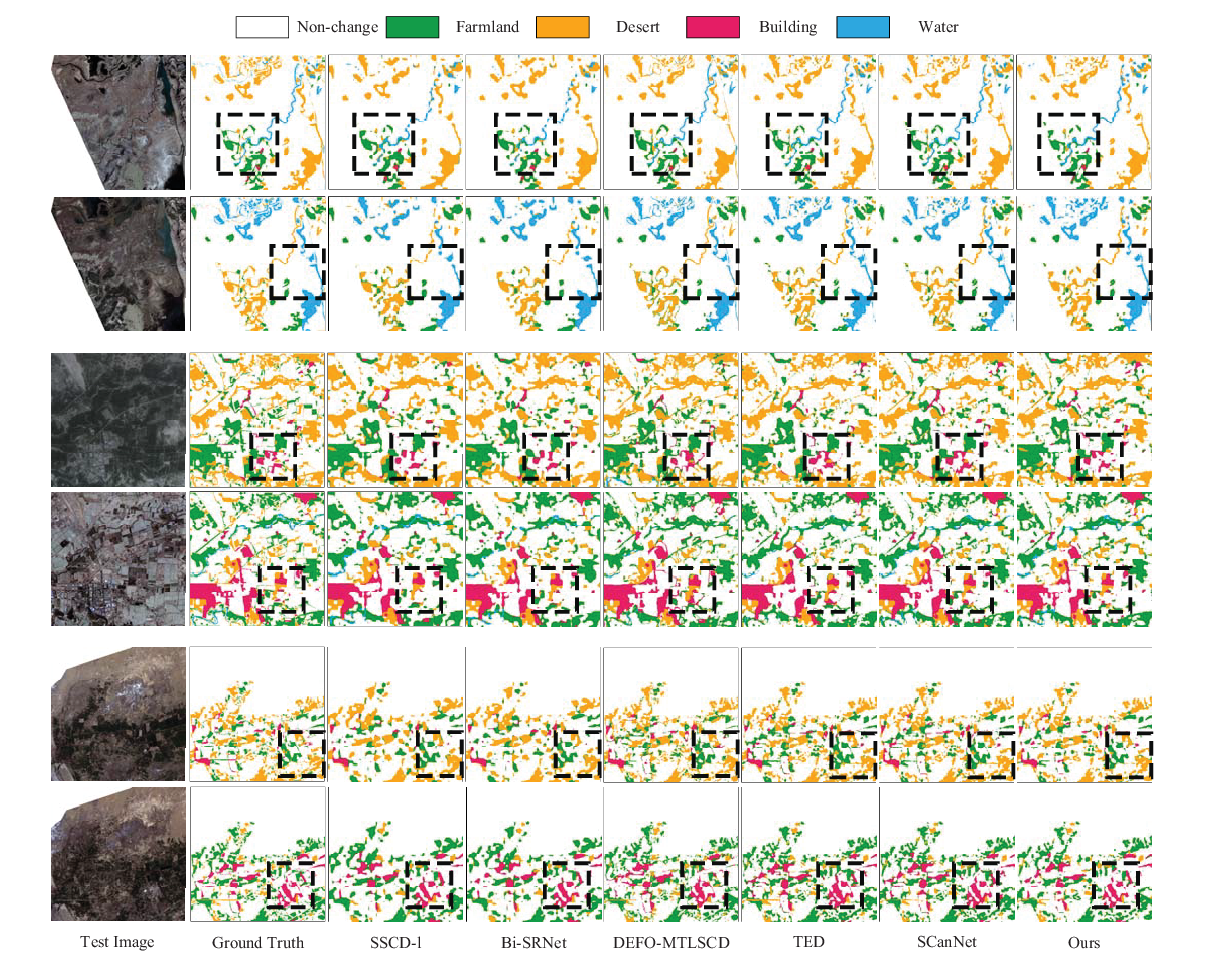}
	\end{tabular}
	\caption{Qualitative experimental results on the Landsat-SCD dataset.}
	\label{visual-landsat-scd}
\end{figure*}

Fig. \ref{visual-second} shows the semantic change detection results on the SECOND dataset produced by several methods. By observing the change maps, it is evident that our proposed method performs exceptionally well on several key categories, including buildings, non-vegetated surfaces, and low vegetation (highlighted with black dashed boxes). Compared to other approaches, our method significantly improves the prediction accuracy for these categories and effectively avoids semantic confusion. For instance, the boundaries between buildings and non-vegetated surfaces are much clearer, and the distinction between low vegetation and trees is delineated more accurately. Overall, our method not only enhances the recognition of changes in these categories but also demonstrates greater robustness and consistency when handling complex semantic classes.

Fig. \ref{visual-landsat-scd} displays the semantic change detection effect maps on the Landsat-SCD dataset obtained by various methods. As shown in Fig. \ref{visual-landsat-scd}, our method exhibits a pronounced advantage in identifying semantic change regions related to farmland, desert, and buildings (indicated by black dashed boxes). In particular, with regard to detecting changes in farmland and buildings, our method achieves higher accuracy, clearly distinguishing subtle differences between agricultural areas and urban structures. Additionally, for the semantic change detection of desert regions and their surroundings, our approach also shows superior performance by accurately capturing detailed changes along the desert boundaries. This demonstrates that, in complex environments, our proposed method has strong adaptability and superiority across multiple categories and scenarios.

In summary, the proposed method, which introduces a multi-task remote sensing image semantic change detection approach guided by graph aggregation prototype learning, achieves significantly superior experimental results compared to existing methods across various complex scenarios. In particular, it demonstrates a strong ability to recognize and differentiate changes in complex ground environments (e.g., vegetation, desert, buildings). Thanks to the effective design of the multi-task joint optimization method, graph aggregation prototype learning, and feature interaction module, the proposed method effectively alleviates conflicts in multi-task learning and improves both the accuracy and robustness of the change detection task, especially when dealing with complex ground environments.

\subsection{Ablation Study}			
To validate the effectiveness of each module in our proposed method, ablation experiments were conducted on the SECOND dataset, which features a larger number of semantic classes. Specifically, when the graph aggregation prototype learning module is removed, the overall loss function becomes ${L_{merge}}$, eliminating the need for the gradient rotation method; when the self-query multi-level feature interaction module is removed, for ease of comparison, the multi-level features are interpolated individually to a common spatial size and then concatenated as input to the classifier; when the bi-temporal feature fusion module is removed, the fusion method is replaced with a simple concatenation operation for comparison; and when the multi-task optimization method is removed, the overall loss function is simply the direct sum of the losses for the three tasks. The experimental results are shown in Table \ref{Ablation-Study}.

From the ablation experiments on the SECOND dataset shown in Table \ref{Ablation-Study}, it can be observed that removing any component of our proposed method results in a certain degree of performance degradation. First, when the graph aggregation prototype learning module is removed, the mIoU drops to 72.19\% compared to 73.03\% for the complete model, and the recognition accuracy of the “low vegetation” category decreases from 56.42\% to 53.55\%. This result indicates that the graph aggregation prototype learning plays a significant role in category-level domain alignment, effectively mitigating the interference of irrelevant changes on change detection, especially in complex scenarios.

Secondly, removing the self-query multi-level feature interaction module has varying impacts on multiple categories. For example, the accuracy of the “trees” category decreases from 45.63\% to 43.03\%, and that of the “playground” category drops from 70.60\% to 66.10\%. This demonstrates that the self-query multi-level feature interaction module helps enhance the expressive power of multi-level features, thereby effectively improving semantic segmentation accuracy. The ablation results for the bi-temporal feature fusion module show that this module is crucial for capturing changes between bi-temporal images. Specifically, the performance of the “low vegetation” category decreases from 56.42\% to 52.15\%, and there is also a slight decline in the “water” category, indicating that the bi-temporal feature fusion module effectively aggregates features from different time points to enhance the model’s recognition ability in dynamic change scenarios.

Thirdly, when the multi-task optimization method is removed, although the overall OA and mIoU do not change significantly, there are some fluctuations in the performance of categories such as “buildings” and “playground,” with the “playground” category dropping from 70.60\% to 65.71\%. This suggests that the multi-task optimization method, through adaptive weight allocation and gradient rotation, successfully alleviates conflicts among different tasks during multi-task learning, thereby further improving the overall stability and performance of the model.
\begin{table*}[htbp]
	\centering
	\renewcommand{\arraystretch}{1.3}
	\caption{Ablation Study Results on the SECOND Dataset.}
	\label{Ablation-Study}
	\resizebox{\textwidth}{!}{
		\begin{tabular}{c|cccc|c}
			\hline
			\textbf{Category} 
			& \textbf{w/o. GAPL} 
			& \textbf{w/o. SQMLFI} 
			& \textbf{w/o. BTFF} 
			& \textbf{w/o. MTO} 
			& \textbf{Ours} \\
			\hline
			
			Non-change     
			& 92.67 & 92.99 & 92.34 & 92.80 & 92.95 \\
			
			Low vegetation 
			& 53.55 & 53.74 & 52.15 & 52.78 & 56.42 \\
			
			N.v.g. surface 
			& 62.52 & 62.17 & 62.23 & 62.94 & 63.03 \\
			
			Trees          
			& 43.03 & 45.63 & 41.79 & 42.04 & 43.28 \\
			
			Water          
			& 54.96 & 57.33 & 56.98 & 58.59 & 57.57 \\
			
			Buildings      
			& 74.47 & 71.08 & 74.11 & 75.12 & 74.62 \\
			
			Playground     
			& 66.10 & 69.61 & 68.53 & 65.71 & 70.60 \\
			
			\hline
			
			OA (\%)        
			& 87.38 & 87.06 & 87.15 & 87.20 & \textbf{87.56} \\
			
			mIoU (\%)      
			& 72.19 & 72.27 & 71.87 & 72.36 & \textbf{73.03} \\
			
			SeK (\%)       
			& 21.65 & 21.61 & 21.91 & 21.72 & \textbf{22.62} \\
			
			${F_{scd}}$ (\%) 
			& 61.83 & 62.04 & 62.08 & 61.94 & \textbf{62.75} \\
			
			\hline
		\end{tabular}
	}
\end{table*}

\begin{figure}[htbp]
	\centering
	\subfloat[w/o. GAPL]{
		\includegraphics[width=0.5\linewidth]{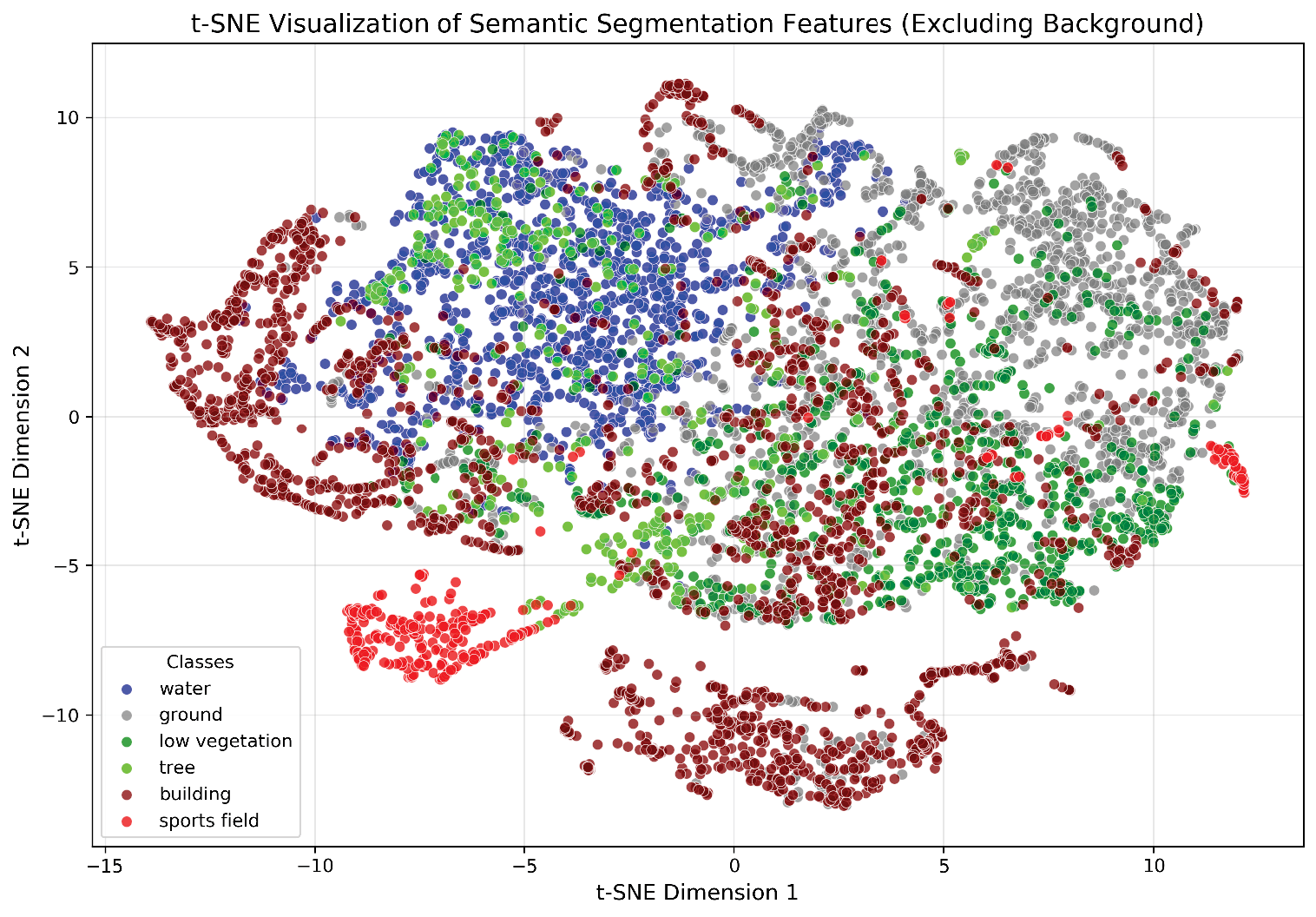}
	}
	\subfloat[w. GAPL]{
		\includegraphics[width=0.5\linewidth]{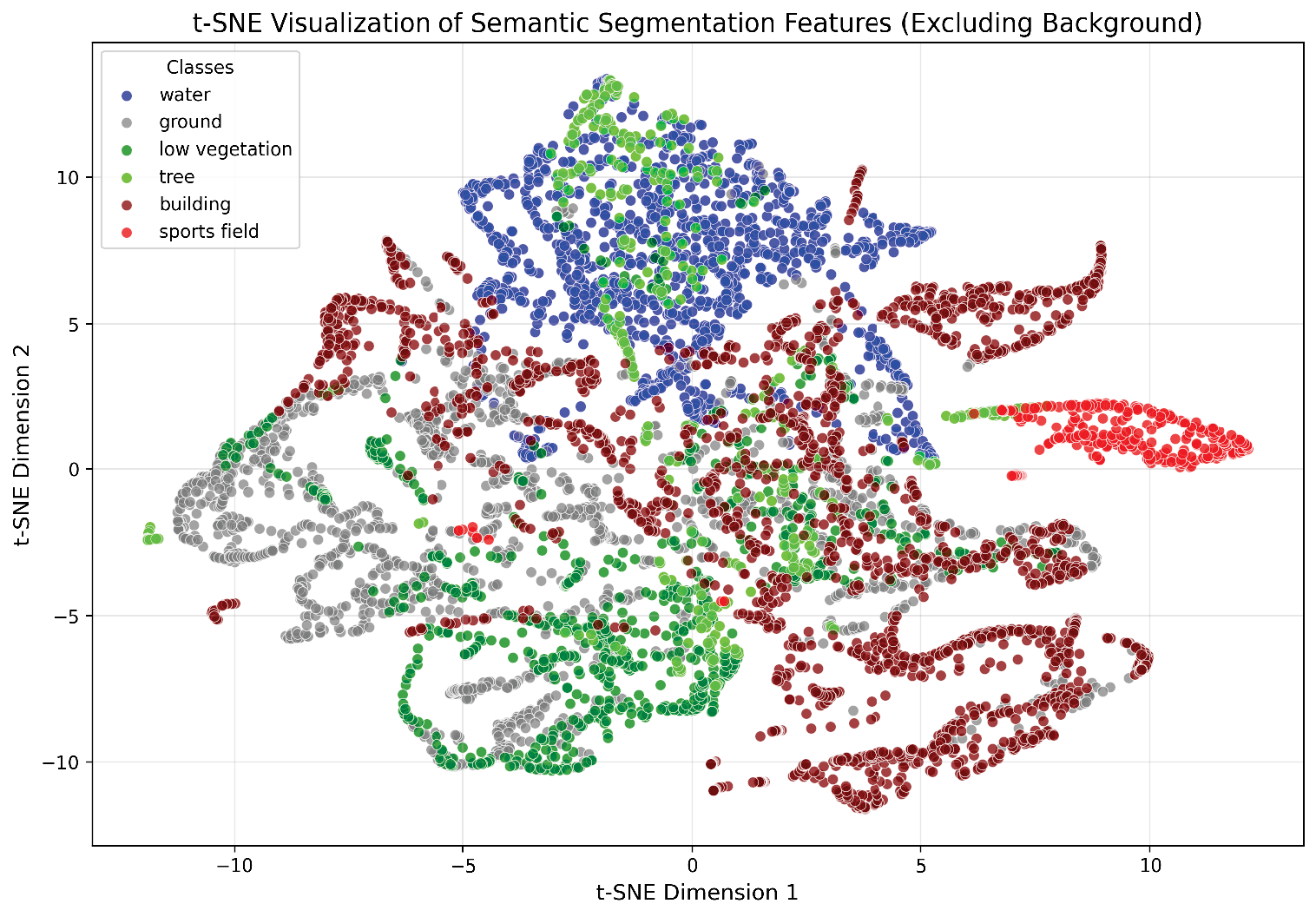}
	}	
	\caption{The t-SNE visualization of feature distributions on the SECOND dataset. (a) Without Graph Aggregation Prototype Learning (GAPL), the feature distributions of different categories are scattered, especially for the "low vegetation" category, which overlaps significantly with other categories. (b) With GAPL, the feature distributions become more compact and well-separated, with reduced overlap between categories, particularly for the "low vegetation" category.}
	\label{t-sne}
\end{figure}

To further visually demonstrate the effectiveness of GAPL, a t-SNE visualization experiment was conducted on the SECOND dataset, as shown in Fig. \ref{t-sne}. By comparing the visualization results of the model without graph aggregation prototype learning and the complete model, the optimization effect of graph aggregation prototype learning on feature distribution can be clearly observed. In the visualization without graph aggregation prototype learning, the feature distributions of different categories are relatively scattered, especially for the "low vegetation" category, which shows significant overlap with other categories. This indicates that irrelevant variations greatly interfere with feature learning, leading to reduced distinguishability between categories. In contrast, in the visualization of the complete model, the feature distributions of each category are more compact and have clearer boundaries. The feature aggregation of the "low vegetation" category is significantly improved, and the overlapping regions with other categories are noticeably reduced.

In summary, these results demonstrate that each module in the proposed method plays an indispensable role in their respective tasks, collectively enhancing the model’s accuracy and adaptability to complex scenarios. In particular, the graph aggregation prototype learning and bi-temporal feature fusion modules are critical for mitigating the impact of irrelevant changes and capturing bi-temporal features.

\section{Conclusions}
This paper proposes a novel semantic change detection framework, GAPL-SCD, to address the challenges of negative transfer and irrelevant changes in bi-temporal remote sensing images. The framework introduces a multi-task joint optimization method that optimizes semantic segmentation, change detection, and an auxiliary graph aggregation prototype learning task. By leveraging adaptive weight allocation and gradient rotation, the proposed method effectively alleviates inter-task conflicts, enhancing multi-task learning capabilities. The graph aggregation prototype learning module constructs an interaction graph based on high-level features, using prototypes as class proxies to perform category-level domain alignment across time points. This significantly reduces the impact of irrelevant changes and other confounding factors on change detection. Additionally, the self-query multi-level feature interaction module and bi-temporal feature fusion module are designed to strengthen multi-scale feature representation, improving the network's ability to handle complex scenarios. Extensive experiments on the SECOND and Landsat-SCD datasets demonstrate that GAPL-SCD effectively optimizes multi-task training, maintains semantic consistency between bi-temporal images, and overcomes the effects of irrelevant changes, achieving state-of-the-art performance in semantic change detection.

In future work, the graph aggregation prototype learning module could be extended to incorporate temporal dependencies beyond bi-temporal images, enabling the framework to handle multi-temporal sequences.

\bibliographystyle{elsarticle-num-names}
\bibliography{myref}

\end{sloppypar}
\end{document}